%% file: main.tex
\documentclass[conference]{IEEEtran}
\IEEEoverridecommandlockouts
\usepackage{cite}
\usepackage{amsmath,amssymb,amsfonts}
\usepackage{algorithmic}
\usepackage{graphicx}
\usepackage{textcomp}
\usepackage{xcolor}
\usepackage{booktabs} 
\usepackage{subcaption}
\usepackage{multirow} 
\usepackage{url} 
\usepackage{tabularx}

\def\BibTeX{{\rm B\kern-.05em{\sc i\kern-.025em b}\kern-.08em
    T\kern-.1667em\lower.7ex\hbox{E}\kern-.125emX}}
\begin{document}

\title{Comprehensive Metapath-based Heterogeneous Graph Transformer for Gene-Disease  Association Prediction
}

\author{\IEEEauthorblockN{Wentao Cui$^{1,2,3}$, Shoubo Li$^{1,2}$, Chen Fang$^{2,4}$, Qingqing Long$^{1,2}$, Chengrui Wang$^{1}$, \\Xuezhi Wang$^{1,2,3*}$ and Yuanchun Zhou$^{1,2,3}$}
\IEEEauthorblockA{$^1$ Computer Network Information Center, Chinese Academy of Sciences, China \\
$^2$ University of Chinese Academy of Sciences, China\\
$^3$ Hangzhou Institute for Advanced Study, University of Chinese Academy of Sciences, China\\
$^4$ State Key Laboratory of Stem Cell and Reproductive Biology, Institute of Zoology, Chinese Academy of Sciences, China\\
\{cuiwentao, sbli, qqlong, crwang, wxz, zyc\}@cnic.cn, fangchen23@ioz.ac.cn}
 \thanks{$*$ Corresponding author.}
}

\maketitle
\input{doc/0_abstract}

\input{doc/1_introduction}
\input{doc/2_Materials_and_methods}
\input{doc/3_experiment}
\input{doc/4_results}
\input{doc/5_discussion}

\section*{Acknowledgment}
This study is supported by grants from the Strategic Priority Research Program of the Chinese Academy of Sciences XDA0460101, the Strategic Priority Research Program of the Chinese Academy of Sciences XDB38030300. 

\bibliographystyle{plain}
\bibliography{ref}
\end{document}

%% file: doc/0_abstract.tex
\begin{abstract}
Discovering gene-disease associations is crucial for understanding disease mechanisms, yet identifying these associations remains challenging due to the time and cost of biological experiments. Computational methods are increasingly vital for efficient and scalable gene-disease association prediction. Graph-based learning models, which leverage node features and network relationships, are commonly employed for biomolecular predictions. However, existing methods often struggle to effectively integrate node features, heterogeneous structures, and semantic information.
To address these challenges, we propose COmprehensive MEtapath-based heterogeneous graph Transformer(COMET) for predicting gene-disease associations. COMET integrates diverse datasets to construct comprehensive heterogeneous networks, initializing node features with BioGPT. We define seven Metapaths and utilize a transformer framework to aggregate Metapath instances, capturing global contexts and long-distance dependencies. Through intra- and inter-metapath aggregation using attention mechanisms, COMET fuses latent vectors from multiple Metapaths to enhance GDA prediction accuracy. Our method demonstrates superior robustness compared to state-of-the-art approaches.
Ablation studies and visualizations validate COMET's effectiveness, providing valuable insights for advancing human health research.
\end{abstract}
\begin{IEEEkeywords}
gene-disease associations, transformer, graph representation learning, meta-path, heterogeneous graph
\end{IEEEkeywords}

%% file: doc/1_introduction.tex
\section{Introduction}
Identifying gene–disease associations (GDAs) is crucial in biomedicine for understanding disease mechanisms \cite{barabasi2011network}. Traditional methods like genetic linkage analysis \cite{ott2015genetic} and genome-wide association studies (GWAS) \cite{manolio2010genomewide} are valuable but can be costly and time-consuming. To expedite the discovery of disease-related genes, new computational models are needed to predict pathogenic genes from existing data \cite{chen2015rbmmmda}. Current approaches for predicting GDAs fall into three main categories: machine learning-based methods, deep learning-based methods, and graph representation learning-based methods.

\textbf{Machine learning-based methods} use algorithms like support vector machines (SVM) and random forests to predict GDAs. Xu et al. \cite{xu2011prioritizing} employed SVM to rank prostate cancer genes, while Chen et al. \cite{chen2018novel} introduced RFMDA, a random forest-based model. However, these approaches often require manual feature extraction, which can be time-consuming and potentially biased. In contrast, \textbf{deep learning-based methods} automatically extract high-level features from raw data, enhancing GDA prediction accuracy and robustness. For instance, RENET \cite{wu2019renet} uses text mining to analyze sentence correlations for GDAs, and PheSeq \cite{yao2024pheseq} applies Bayesian deep learning to integrate phenotype data. Despite their potential, these methods require substantial training data and often lack interpretability due to their black-box nature.

With the rapid growth of modern biotechnology, extensive biological network data has driven the development of network biology and network medicine \cite{meng2022weighted}. Researchers are integrating diverse datasets to build heterogeneous biological networks for predicting GDA using \textbf{graph representation learning based-methods}. It is formulated as a link prediction task, where relationships between diseases and genes can be learned from different dimensions of heterogeneous networks. Graph representation learning methods can be categorized into three types based on how they obtain node representations. The first type is \textbf{random walk-based algorithms} on graphs. Valdeolivas et al. \cite{valdeolivas2019random} constructed gene-protein and disease-disease networks, using a restart random walk method to predict disease-related candidate genes. However, the node sequences generated by random walks start locally and do not directly leverage the constructed graph's global structural information. As a result, they may fail to capture global consistency patterns and features comprehensively, thereby limiting their ability to fully express complex relationships between nodes. 
In contrast, the second type involves \textbf{GNN-based methods} that leverage global consistency to model complex relationships within the graph. Rao et al. \cite{rao2018phenotype} incorporated pathway and phenotype data into disease gene networks using graph convolution methods to identify potential disease genes. 
Xiang et al. \cite{xiang2021prgefne} utilized phenotype and gene ontology (GO) information to build a heterogeneous network, applying fast network embedding to predict disease-related genes. However, these methods do not effectively leverage prior biological knowledge about known GDA.
The third type involves \textbf{metapath-based methods}, which construct metapaths based on prior knowledge to obtain node representations. Renet utilized five meta paths in the heterogeneous disease-gene network, employing singular value decomposition to create low-dimensional feature matrices for diseases and genes, facilitating disease-gene association prediction \cite{wu2019renet}. Despite recent advances in gene-disease association prediction, several limitations persist:
\begin{itemize}
    \item \textbf{Limited exploration in GDA mining.} Current methods \cite{yang2018hergepred,liu2021pathogenic} do not fully capture complex relationships in bioinformatics networks, potentially overlooking important gene–disease associations critical for understanding diseases and developing treatments.

    \item \textbf{Ignoring long-distance structural correlations.} Most approaches use mean pooling, which smooths out differences among distant yet functionally similar genes. This can hide important interactions, reducing the accuracy and reliability of predictions.

    \item \textbf{Neglect of characteristic node information.} Many methods initialize gene and disease nodes with random or generic values, missing key characteristics that shape biological context and impact disease association predictions.
\end{itemize}


To tackle the aforementioned limitations, we proposed a \textbf{CO}mprehensive \textbf{ME}tapath-based heterogeneous graph \textbf{T}ransformer (COMET) for GDAs.
Firstly, we construct heterogeneous networks for diseases and genes using diverse types of data including diseases, genes, ontologies, and phenotypes. We initialize node vectors in the graph using BioGPT \cite{luo2022biogpt}, a domain-specific language representation model pretrained on large-scale biomedical corpora. Next, we define seven Metapaths and employ 
transformer \cite{valdeolivas2019random} to aggregate Metapath instances, capturing global contexts to model long-distance dependencies. Finally, we conduct intra-metapath aggregation and inter-metapath using the attention mechanism to fuse latent vectors obtained from multiple metapaths into final node embeddings for gene-disease association prediction.

Overall, our contributions to this work can be summarized as follows:
\begin{itemize}
    \item  We propose COMET, a framework designed to capture comprehensive structural and semantic information across diseases, genes, ontologies, and phenotypes for GDA prediction.
    \item 
    To the best of our knowledge, we are the first to explore the pre-trained language models for predicting GDA.
    \item COMET introduces a transformer to message passing between the nodes of different types, and adopts a two-level attention mechanism to uncover crucial biological relationships, which outperforms various state-of-the-art gene-disease prediction methods.
\end{itemize}

%% file: doc/2_Materials_and_methods.tex
\section{Materials and methods}

\begin{figure*}[t!]
\centering
\includegraphics[width=1.0\linewidth]{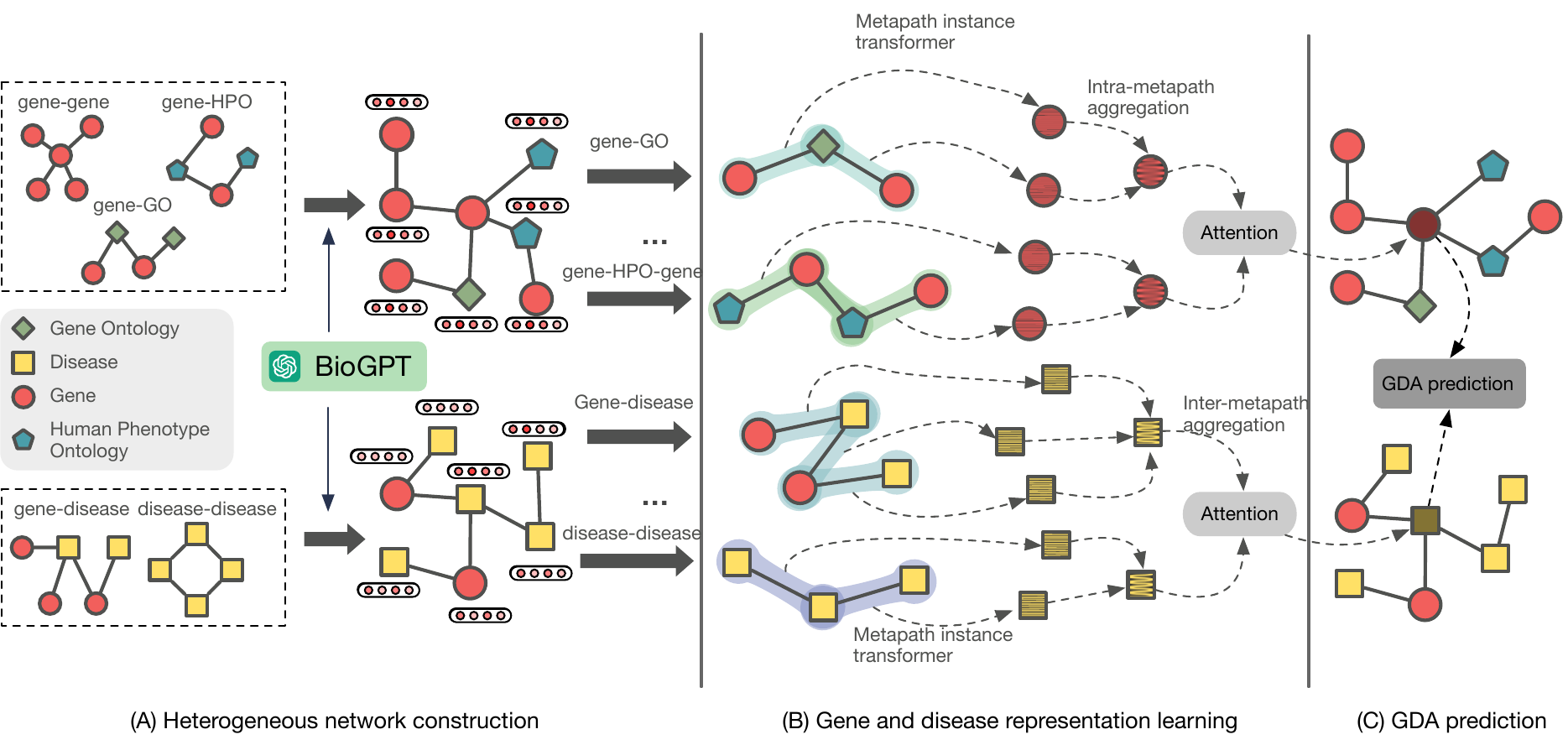}
\caption{Overview of the proposed COMET.(A) Heterogeneous network construction:We constructed heterogeneous graphs of genes and diseases from five subnetworks and initialized the node feature using BioGPT. (B) Gene and disease representation learning: We define seven Metapaths and use a transformer to aggregate them, and then we employ intra- and inter-metapath aggregation with the attention mechanism to fuse latent vectors. (C) GDA prediction: } \label{fig:framework}
\end{figure*}

In this section, we will introduce the proposed COMET. The overall model consists of three parts: heterogeneous network construction, gene and disease representation learning and GDA prediction. More information can be found in Figure. \ref{fig:framework}.


\subsection{Preliminary}
In this section, we give formal definitions of some important terminologies related to heterogeneous graphs. 

\noindent\textbf{Definition 1 (\textit{Heterogeneous Graph}).} A heterogeneous graph \( G = (V, E) \) is defined as a graph associated with a node type mapping function \( \varphi : V \rightarrow A \) and an edge type mapping function \( \psi : E \rightarrow B \). \( A \) and \( B \) denote the predefined sets of node types and edge types, respectively, with \(|A| + |B| > 2\).

\noindent\textbf{Definition 2 (\textit{Metapath}).} A metapath \( R \) is defined as a path in the form of 
\[ 
A_1 \xrightarrow{C_1} A_2 \xrightarrow{C_2} \cdots \xrightarrow{C_l} A_{l+1} 
\]
(abbreviated as \( C_1 C_2 \cdots C_{l+1} \)), which describes a composite relation \(C = C_1 \circ C_2 \circ \cdots \circ C_l \) between node types \( A_1 \) and \( A_{l+1} \), where \( \circ \) denotes the composition operator on relations.

\noindent\textbf{Definition 3 (\textit{Metapath Instance}).} Given a metapath \( P \) of a heterogeneous graph, a metapath instance \( R_{(i,j)} \) of \( R \) is  defined as a node sequence in the graph G following the schema defined by \( R \) with start node i and end node j.

\noindent\textbf{Definition 4 (\textit{Metapath-based Neighbors)}.} Given a metapath \( R \) of a heterogeneous graph, the metapath-based neighbors \( N_{v}^{R} \) of a node \( v \) is defined as the set of nodes that connect with node \( v \) via metapath instances of \( R \). A neighbor connected by two different metapath instances is regarded as two different nodes in \( N_{v}^{R} \). Note that \( N_{v}^{R} \) includes \( v \) itself if \( R \) is symmetric.

\subsection{Heterogeneous network construction}
The pathogenesis of diseases is influenced by factors like environment and genetics \cite{li2021evaluating}. To understand the complex relationships between genes and diseases, we integrate diverse data sources—phenotype, protein, and RNA—into a heterogeneous network model using five resources: HumanNet \cite{hwang2019humannet}, Human Phenotype Ontology (HPO) \cite{robinson2008human}, Gene Ontology (GO) \cite{ashburner2000gene}, Disease Ontology (DO) \cite{schriml2012disease}, and DisGeNet \cite{pinero2016disgenet}.

\textbf{Gene Heterogeneous Network.} We construct a gene heterogeneous network by combining gene–gene, gene–GO, and gene–HPO associations. Gene–gene interactions, derived from HumanNet, are represented by a similarity score \(S(g_i, g_j)\):

\[
S_{\text{gene-gene}}[i][j] = 
\begin{cases}
1, & \text{if } e_{i,j} \text{ exists in HumanNet} \\
0, & \text{otherwise}
\end{cases}
\]

The gene–GO and gene–HPO networks are similarly constructed using data from GO and HPO, respectively, resulting in a network with five interaction types: gene–gene, gene–HPO, gene–GO, HPO–gene, and GO–gene.

\textbf{Disease Heterogeneous Network.} We use gene–disease associations from DisGeNet and disease–disease associations from Disease Ontology to form a disease network with three types of interactions: disease–disease, disease–gene, and gene–disease.

\textbf{Node Feature Initialization.} Node features for genes and diseases are initialized with BioGPT \cite{luo2022biogpt}, which provides contextual representations based on gene and disease names. This initialization improves the model’s ability to capture complex relationships and enhance predictions of gene–disease associations.

\subsection{Gene and disease representation learning }
\textbf{Metapath instance transformer.} We construct seven types of metapath instances: gene-gene (g-g), gene–HPO–gene(g-h-g), gene-Go-gene (g-o-g), gene-disease (g-d), disease-disease(d-d), gene-disease-gene (g-d-g) and disease-gene-disease (d-g-d). In bioinformatics, heterogeneous networks pose a challenge as different node types have distinct feature spaces. To address this, we transform features of various node types into a unified latent vector space, obtaining the initial embedding \( \mathbf{h}_i \) for each node \( i \). Capturing a node's hidden features involves modeling its context through metapath instances. A metapath instance-level transformer maps these sequences into continuous representations and averages them to obtain the metapath instance embedding. Specifically, the transformer converts each metapath instance sequence \( R_k(i, j) \) into the metapath instance embedding \( \mathbf{Z}_{R_k(i,j)} \) as follows:

\begin{equation}
\mathbf{Z}_{R_k(i,j)} = \text{Transformer}\left( \big\| (\mathbf{h}_t \mid \forall t \in R_k(i,j)) \right),
\end{equation}
The metapath transformer encoder features multi-head attention and a feed-forward network. It processes queries, keys, and values in parallel, incorporating residual connections, normalization, and ReLU activation. Node positions within metapaths offer key information for predicting gene-disease associations, particularly distant nodes with pharmacological relevance. By combining multi-head self-attention with positional embeddings, it captures long-range dependencies and improves gene-disease prediction performance significantly.



\textbf{Intra-metapath aggregation} The target node \( i \) possesses multiple instances of certain types of metapaths. The idea of semantic aggregation for single-pathway semantics is to aggregate the semantic information of multiple instances of the same metapath type into the corresponding target node. Inspired by the work of Wang et al. \cite{wang2019heterogeneous}, we adopt a self-attention mechanism to aggregate the metapath instance information of the target node. In a heterogeneous network, the target node \( i \) has multiple instances of certain types of metapaths, but the importance of different metapath instances varies. Therefore, we introduce the weight \( \alpha_{ij}^R \) for the target node \( i \), which measures the importance of the target node \( i \). The weight \( w_{ij}^R \) of metapath instance \( R_k(i, j) \) and the weight \( w_{ii}^R \) of the target node \( i \) can be described as follows:

\begin{equation}
w_{ij}^{R_{k}} = \text{Attention}(h_i'; Z_{R_k(i,j)}'; R_k),
\end{equation}

\begin{equation}
w_{ii}^{R_{k}} = \text{Attention}(h_i'; R_k),
\end{equation}
in this paper, we use the softmax function to normalize the weight coefficients. For \( i \in N_v^R \), the normalized weight coefficients \( \alpha_{ij}^R \) and \( \alpha_{ii}^R \) can be uniformly calculated:

\begin{equation}
\alpha_{im}^{R_{k}} = \frac{\exp(w_{im}^{R_{k}})}{\sum_{n \in N_i^R} \exp(w_{in}^{R_{k}})},
\end{equation}
this process ensures that the weights of different metapath instances and the target node are properly weighted for subsequent aggregation.

To address the high variance issue in network data, we extends single-semantic attention to multi-head single-semantic attention to achieve stable model training. Given a metapath type \( R_k \) and a target node \( i \), the embedding $Z_i^{R_{k}}$  of the target node under specific metapath \( R_k \) semantics can be learned by repeating \( H \) attention calculations and concatenating the embedding results, as shown in:

\begin{equation}
Z_i^{R_{k}} = \prod_{h=1}^{H} \sigma \left( \sum_{j \in N_i^{R_k-i}} \left[ \alpha_{ij}^{R_{k}} \right]_h \cdot h^{'}_{R_{k}(j)} + \left[ \alpha_{ii}^{R_k} \right]_h \cdot h_i' \right),
\end{equation}
here, \( \sigma(\cdot) \) denotes the activation function, \( \left[ \alpha_{ij}^{R_k} \right]_h \) represents the normalized importance of node \( i \) for metapath instance \( R_k(i, j) \) at attention head \( H \), and \( \left[ \alpha_{ii}^{R_k} \right]_h \) represents the normalized importance of central node \( i \) at attention head \( H \).

\textbf{Inter-metapath aggregation.}
Different metapath types have distinct bioinformatics meanings. To learn the semantic representation of a target node across multiple metapath types, semantic aggregation of multiple metapath types is required. Similar to intra-metapath aggregation, different metapath types also have varying importance for the central node. For example, different metapath types, such as d-d and d-g-d, exhibit significant differences in the importance of embedding for disease nodes. The differences in metapath types need to be distinguished and reflected in the multi-metapath semantic aggregation of the target node. Therefore, after aggregating node and edge information for each metapath type, this paper employs a multi-metapath semantic attention layer to combine semantic information from all metapath types. This paper introduces the weight \( \beta_{R_k} \) for each metapath type \( R_k \) to describe the differences in importance for the target node \( i \):

\begin{equation}
\beta_{R_k} = \text{Attention}(Z_i^{R_k}),
\end{equation}
then, the normalized weight coefficients \( \beta_{R_k} \) are obtained through the softmax function.

\begin{equation}
\beta_{R_k} = \frac{\exp(w_{R_k})}{\sum_{R_t \in \mathcal{R}} \exp(w_{R_t})},
\end{equation}
given the target node \( i \), its final representation vector \( Z_i \) can be obtained using the aggregated representation vectors \( Z_i^{R_k} \) and normalized weight coefficients \( \beta_{R_k} \):

\begin{equation}
Z_i = \sum_{R_k \in \mathcal{R}} \beta_{R_k} \cdot Z_i^{R_k},
\end{equation}
here, \( \mathcal{R} \) represents the set of all metapaths.

\subsection{GDA prediction}
After obtaining the low-dimensional multi-semantic representations of genes and diseases, we project the node embeddings into a space with node similarity measures for the downstream gene-disease prediction task. Given the gene embedding \( \mathbf{z}_g \) and the disease embedding \( \mathbf{z}_d \), we optimize the model weights by minimizing the following loss function through negative sampling:

\begin{equation}
L = - \sum_{(d,g) \in \mathcal{D}} \log \sigma (\mathbf{z}_d^T \cdot \mathbf{z}_g) - \sum_{(d',g') \in \mathcal{D}^-} \log \sigma (-\mathbf{z}_{d'}^T \cdot \mathbf{z}_{g'}),
\end{equation}
where \( \sigma(\cdot) \) is the sigmoid function, \( \sigma(\mathbf{z}_d^T \cdot \mathbf{z}_g) \) is the probability that gene \( g \) interacts with disease \( d \), \( \mathcal{D} \) is the set of observed node pairs, and \( \mathcal{D}^- \) (the complement of \( \mathcal{D} \)) is the set of non-existing gene-disease pairs sampled from all unobserved gene-disease pairs.

%% file: doc/3_experiment.tex
\section{Experiment}

\subsection{Datasets}
In this section, we detail our dataset, constructed from five public sources: HumanNet, GO, HPO, DO, and DisGeNet, as outlined in Table \ref{tab:datasets}. \textbf{HumanNet} provides a human gene network, including 18,462 genes and their functional associations through protein-protein interactions, mRNA co-expression, and genomic context. \textbf{GO (Gene Ontology)} describes gene functions with unique identifiers, linking genes to multiple terms, aiding in reconstructing the gene heterogeneous network. \textbf{HPO (Human Phenotype Ontology)} outlines relationships between phenotypic abnormalities in diseases, offering insights into gene functions and genetic networks. \textbf{Disease Ontology (DO)} provides a standardized classification and clear definitions for diseases, which we use to calculate disease similarities. \textbf{DisGeNet} compiles information on genes and mutations associated with diseases from various sources, supporting research on disease genes and validation of predictions.


\begin{table}[h!]
    \centering
    \caption{Statistics of datasets}
    \resizebox{0.48\textwidth}{!}{%
    \begin{tabular}{l l l l l l}
        \toprule
        \textbf{Database} & \textbf{\# Nodes}  & \textbf{\# Edges}&\textbf{Relations} \\
        \midrule
        HumanNet & 18,462 & 1,051,038 & Gene-Gene \\
        Gene Ontology & 19,661  & 290,214 & Gene-Go terms\\
        Human Phenotype Ontology & 16,870  & 182,144 & Gene-Phenotype\\
        Disease Ontology & 6,453  & 13,444 & Disease-Disease \\
        DisGeNet & 21,354  & 86,297 & Gene-Disease \\
        \bottomrule
    \end{tabular}
    }
    \label{tab:datasets}
\end{table}

\subsection{Baselines}
To evaluate the efficacy of our approach, we benchmark it against two random walk-based approaches including HerGePred \cite{yang2018hergepred},
dgn2vec \cite{liu2021pathogenic}, two GCN-based methods including  GCN \cite{kipf2016semi}, HOGCN \cite{kishan2021predicting}, and two metapath-based methods including HNEEM \cite{wang2019predicting}, DGP-PGTN \cite{li2023end}. The detailed introduction to each method can be found as follows.
\begin{itemize}
    \item \textbf{HerGePred} uses random walks for low-dimensional representations of diseases and genes in heterogeneous networks, facilitating association predictions.
    \item \textbf{dgn2vec} employs a random walk algorithm to generate node embeddings, focusing on disease-gene association prediction.
    \item \textbf{GCN} applies a semi-supervised Graph Convolutional Network (GCN) directly to disease and gene heterogeneous networks using the original adjacency matrix.
    \item \textbf{HOGCN} collects neighbor nodes at varying distances to capture informative representations for disease-gene association predictions.
    \item \textbf{HNEEM} extracts node features from multiple heterogeneous networks, concatenating disease and gene vectors for disease-gene pair representations, and uses random forests for classification.
    \item \textbf{DGP-PGTN} integrates diverse data sources and utilizes a parallel graph transformer network to predict GDAs, focusing on capturing latent interactions.
    
\end{itemize}

For all the baselines mentioned above, we conduct experiments
on the same dataset with the same parameters as their experimental settings. We compare COMET with baselines using evaluation measures including AUC, AUPR, precision, recall, and F1-measure, with AUC and AUPR as primary metrics for overall performance evaluation, while employing Youden’s index to determine the optimal threshold for calculating accuracy, precision, recall, and F1-score.



%% file: doc/4_results.tex
\section{Results}

\subsection{The results of gene-disease prediction}

In our experimental evaluation, we conducted a comprehensive comparison of several state-of-the-art models, aiming to assess their effectiveness in predicting GDAs. As shown in table \ref{tab:comparison-results}, Metapath2vec and HerGePred, based on random walk methods, achieved the poorest performance because they cannot effectively capture the complex relationships present in heterogeneous networks. HNEEM and DGP-PGTN performed better due to the use of pre-defined Metapaths based on prior knowledge, which helps in obtaining better node representations.
Our method, COMET, achieved the best performance. This is primarily because we extracted more Metapath information based on biological prior knowledge and used transformers to capture more comprehensive and context-aware associations. This enhanced semantic aggregation capability allows COMET to outperform other models across various evaluation metrics.
Overall, these results affirm the effectiveness of our proposed approach, showcasing its ability to provide more accurate and reliable predictions for gene-disease associations compared to existing state-of-the-art methods.

\begin{table}[h!]
    \centering
    \caption{Comparison Experiment Results}
     \resizebox{0.48\textwidth}{!}{%
      \begin{tabular}{c c c c c c}
        \toprule
        \textbf{Model} & \textbf{Precision} & \textbf{Recall} & \textbf{F1}  & \textbf{AUC} & \textbf{AUPR} \\
        \midrule
        dgn2vec & 0.8504 & 0.8454 & 0.8445 & 0.8799 & 0.8428\\
        HerGePred & 0.8749 & 0.8747 & 0.8746 & 0.9291 & 0.9015 \\ 
        \midrule
        GCN &   0.9112 & 0.9119 & 0.9120  & 0.9417 & 0.8991 \\
        HOGCN & 0.9447 & 0.9406 & 0.9426  & 0.9582 & 0.9014 \\
        \midrule
         HNEEM & 0.9468 & 0.9469 & 0.9468  & 0.9647 & 0.9273 \\
        DGP-PGTN & 0.9527 & 0.9483 & 0.9505  & 0.9650 & 0.9315 \\
        \midrule
       \textbf{COMET} & \textbf{0.9693} & \textbf{0.9679} & \textbf{0.9682}  & \textbf{0.9811} & \textbf{0.9648} \\
        \bottomrule
    \end{tabular}}
    \label{tab:comparison-results}
\end{table}


\subsection{Ablation Study}
COMET’s strong performance stems from three key innovations: heterogeneous graph node initialization, metapath design based on prior knowledge, and a metapath-based hierarchical transformer and attention network. We performed ablation studies to assess how these components contribute to overall performance.

\textbf{Effects of different node feature.} To assess the impact of different node feature initialization methods on COMET’s performance, we compared random initialization with BioGPT. As shown in Table \ref{tab:CombinedAblation}, initializing node features with BioBERT significantly enhances performance across all metrics compared to random initialization. BioGPT captures relevant semantic information from the biomedical domain, aiding more accurate GDA predictions. Therefore, using BioBERT for node feature initialization improves the model’s ability to learn meaningful representations, leading to better GDA prediction performance.

\textbf{Effects of different metapath type.} To assess the importance of different metapath types in semantic information aggregation, we conducted an ablation study on the disease-gene heterogeneous network. The COMET model was trained using metapaths g-g, g-d-g, g-o-g, d-g-d, g-d, g-h-g, and d-d. As shown in Table \ref{tab:CombinedAblation}, using all metapaths significantly improved performance. Removing the G-D-G and D-G-D metapaths resulted in the largest decline in AUROC and AUPR, as these paths contain essential association information between diseases and genes. The removal of G-O-G and D-S-D also caused a 1\% to 2\% performance drop, indicating that including GO and disease nodes enhances the model’s semantic aggregation capability.

\textbf{Effects of different metapath instance encoders.} In COMET, we use a transformer encoder as a metapath instance aggregator. To evaluate its impact, we conducted ablation experiments with Average pooling, Max-Pooling, Bi-LSTM, Bi-GRU, and the Transformer aggregator. As shown in Table \ref{tab:CombinedAblation}, the transformer aggregator achieved the best performance. Bi-LSTM and Bi-GRU also performed well, as they capture sequential features effectively. In contrast, Average and Max-Pooling led to information loss due to their simple pooling approach.

\begin{table}[h!]
    \centering
    \tiny
    \caption{Results of Ablation Study}
   
    \begin{tabular}{l c c c c c c}
        \toprule
        \textbf{Ablation} & \textbf{Model} & \textbf{Precision} & \textbf{Recall} & \textbf{F1} & \textbf{AUC} & \textbf{AUPR} \\
        \midrule
        \multirow{1}{*}{Node feature} & Random & 0.9331 & 0.9283 & 0.9307 & 0.9503 & 0.9340 \\
        \midrule
        \multirow{4}{*}{\shortstack{Metapath\\Instance\\encoders}} 
        & Average & 0.9427 & 0.9325 & 0.9376 & 0.9514 & 0.9331 \\
        & Max-Pooling & 0.9312 & 0.9378 & 0.9345 & 0.9465 & 0.9337 \\
        & Bi-LSTM & 0.9298 & 0.9332 & 0.9315 & 0.9441 & 0.9334 \\
        & Bi-GRU & 0.9550 & 0.9578 & 0.9564 & 0.9583 & 0.9516 \\
        \midrule
        \multirow{7}{*}{\shortstack{Metapath\\Type}} 
        & w/o g-g & 0.9295 & 0.9385 & 0.9339 & 0.9431 & 0.9354 \\
        & w/o g-d-g & 0.9398 & 0.9327 & 0.9362 & 0.9442 & 0.9345 \\
        & w/o g-o-g & 0.9334 & 0.9350 & 0.9342 & 0.9524 & 0.9450 \\
        & w/o d-g-d & 0.9281 & 0.9335 & 0.9308 & 0.9437 & 0.9325 \\
        & w/o d-d & 0.9410 & 0.9315 & 0.9362 & 0.9446 & 0.9348 \\
        & w/o g-h-g & 0.9469 & 0.9447 & 0.9458 & 0.9511 & 0.9434 \\
        & w/o g-d & 0.9289 & 0.9297 & 0.9293 & 0.9508 & 0.9431 \\
        \midrule
        \multirow{1}{*}{Full model} & COMET & \textbf{0.9693} & \textbf{0.9679} & \textbf{0.9682} & \textbf{0.9811} & \textbf{0.9670} \\
        \bottomrule
    \end{tabular}
    \label{tab:CombinedAblation}
\end{table}

%% file: doc/5_discussion.tex
\subsection{Conclusion}

In this paper, we propose COMET, a comprehensive metapath-based heterogeneous graph
transformer for GDAs prediction. We leverage diverse sources of gene and disease-related information to construct a heterogeneous bioinformatics network. Node features in this network are initialized using pretrained models from the biomedical domain. To capture rich structural and semantic information, we employ a transformer encoder and a two-level attention mechanism. This approach enables us to derive low-dimensional vector representations of genes and diseases for association prediction.
We validate COMET's effectiveness and robustness by comparing it against six state-of-the-art methods using a constructed meta-path dataset for gene-disease prediction. Results show significant performance improvements with COMET. Ablation studies analyze the contributions of different meta-path types and components within COMET to prediction performance. COMET offers a powerful and robust tool for advancing gene-disease association prediction in life sciences. Future work will focus on expanding COMET by integrating additional heterogeneous attribute information and validating prediction results through wet-lab experiments.